\documentclass[letterpaper, 10 pt, conference]{ieeeconf}  

\IEEEoverridecommandlockouts                              

\usepackage[utf8]{inputenc}
\usepackage{graphics} 
\usepackage{epsfig} 
\usepackage{mathptmx} 
\usepackage{amsmath,amssymb,amsfonts}
\usepackage{mathtools}
\usepackage{algorithm}
\usepackage{multirow}
\usepackage{booktabs}
\usepackage{algpseudocode}
\usepackage{tikz}
\usetikzlibrary{arrows.meta, positioning}
\usepackage{bm}
\usepackage{subcaption}
\usepackage{colortbl}
\usepackage{makecell}
\usepackage[nospace,compress, sort]{cite}

\usepackage{hyperref}
\usepackage{siunitx}

\DeclareMathAlphabet{\mathcal}{OMS}{cmsy}{m}{n}

\renewcommand{\vec}{\mathbf}
\newcommand{\set}[1]{\mathbb{#1}}
\newcommand{\R}{\mathbb{R}}
\newcommand{\N}{\mathbb{N}}
\newcommand{\SO}{SO(3)}
\newcommand{\SE}{SE(3)}
\newcommand{\safeset}{\mathcal{SS}}
\newcommand{\cmd}{\text{cmd}}
\newcommand{\co}{\text{c}}
\newcommand{\s}{\text{s}}

\title{\LARGE \bf
Improving Drone Racing Performance Through Iterative Learning MPC
}

\author{Haocheng Zhao, Niklas Schl{\"u}ter, Lukas Brunke, and Angela P. Schoellig
\thanks{The authors are with the Learning Systems and Robotics lab
(\href{www.learnsyslab.org}{www.learnsyslab.org}) at the Technical University of Munich, Germany.
Emails: \{{\tt\small haocheng.zhao}, {\tt\small niklas.schlueter}, {\tt\small lukas.brunke},  {\tt\small angela.schoellig}\}{\tt\small @tum.de}}}

\begin{document}

\maketitle
\thispagestyle{empty}
\pagestyle{empty}

\begin{abstract}
Autonomous drone racing presents a challenging control problem, requiring real-time decision-making and robust handling of nonlinear system dynamics. While iterative learning model predictive control~(LMPC) offers a promising framework for iterative performance improvement, its direct application to drone racing faces challenges like real-time compatibility or the trade-off between time-optimal and safe traversal. In this paper, we enhance LMPC with three key innovations:~(1) an adaptive cost function that dynamically weights time-optimal tracking against centerline adherence,~(2)~a shifted local safe set to prevent excessive shortcutting and enable more robust iterative updates, and~(3) a Cartesian-based formulation that accommodates safety constraints without the singularities or integration errors associated with Frenet-frame transformations. Results from extensive simulation and real-world experiments demonstrate that our improved algorithm can optimize initial trajectories generated by a wide range of controllers with varying levels of tuning for a maximum improvement in lap time by 60.85\%. Even applied to the most aggressively tuned state-of-the-art model-based controller, MPCC++, on a real drone, a 6.05\% improvement is still achieved. Overall, the proposed method pushes the drone toward faster traversal and avoids collisions in simulation and real-world experiments, making it a practical solution to improve the peak performance of drone racing.
\end{abstract}

\section{INTRODUCTION}

Autonomous racing has emerged as a challenging benchmark for high-speed robotics, demanding precise trajectory optimization, real-time decision-making, and robust control under dynamic constraints while operating at the physical limits of the system. Moreover, advances in this area have far-reaching implications, as the underlying principles readily translate to other time-critical applications such as drone-based logistics or emergency response missions~\cite{shakhatreh2019unmanned, lyu2023unmanned, rejeb2023drones}. These applications share the fundamental challenge of achieving aggressive yet safe motion, making racing scenarios an ideal testbed for advancing real-world high-speed robot control. Compared to other racing domains, drone racing is particularly demanding due to its full 3D motion and underactuated nature, requiring control algorithms to operate with extreme precision and robustness while handling the system at the edge of its physical capabilities~\cite{moon2019challenges, foehn2022alphapilot,hanover2024autonomous}.

Current state-of-the-art approaches for drone racing can be broadly categorized into two main groups: Model-based control algorithms and reinforcement learning~(RL)-based methods~\cite{song2023reaching}. Among RL methods, proximal policy optimization~(PPO) has demonstrated state-of-the-art performance but suffers from poor sample efficiency and a high sensitivity to sim-to-real transfer. In contrast, model-based control methods like model predictive contouring control~(MPCC) and its improved variant, MPCC++, offer more consistent performance across repeated trials in real-world scenarios with comparable lap time~\cite{krinner2024mpcc++}. 

Recent advances have led to the development of learning-based control approaches that leverage past trajectory data to iteratively enhance control performance~\cite{brunke2022safe}.
Among these, iterative learning model predictive control~(LMPC) has emerged as a framework that integrates model predictive control~(MPC) with iterative learning control~(ILC), enabling data-driven performance improvements over multiple iterations while combining the strengths of both approaches~\cite{rosolia2017learning}. In this work, we apply the LMPC framework to the drone racing domain. The contribution of this paper is threefold:

\begin{itemize}
\item We adapt the cost function design for LMPC that employs adaptive weight allocation to balance time-optimal tracking with centerline adherence, ensuring an effective trade-off between lap time and gate traversal. 
\item We propose a modification to the local safe set by shifting the safe set across the center line to prevent overly aggressive shortcuts.
\item We present a framework that enhances the LMPC formulation in the Cartesian frame for the drone racing problem. This allows the definition of safety constraints without singularities and numerical integration errors caused by Frenet frame transformations.
\end{itemize}

\begin{figure}[t]
  \centering
  \includegraphics[width=\linewidth]{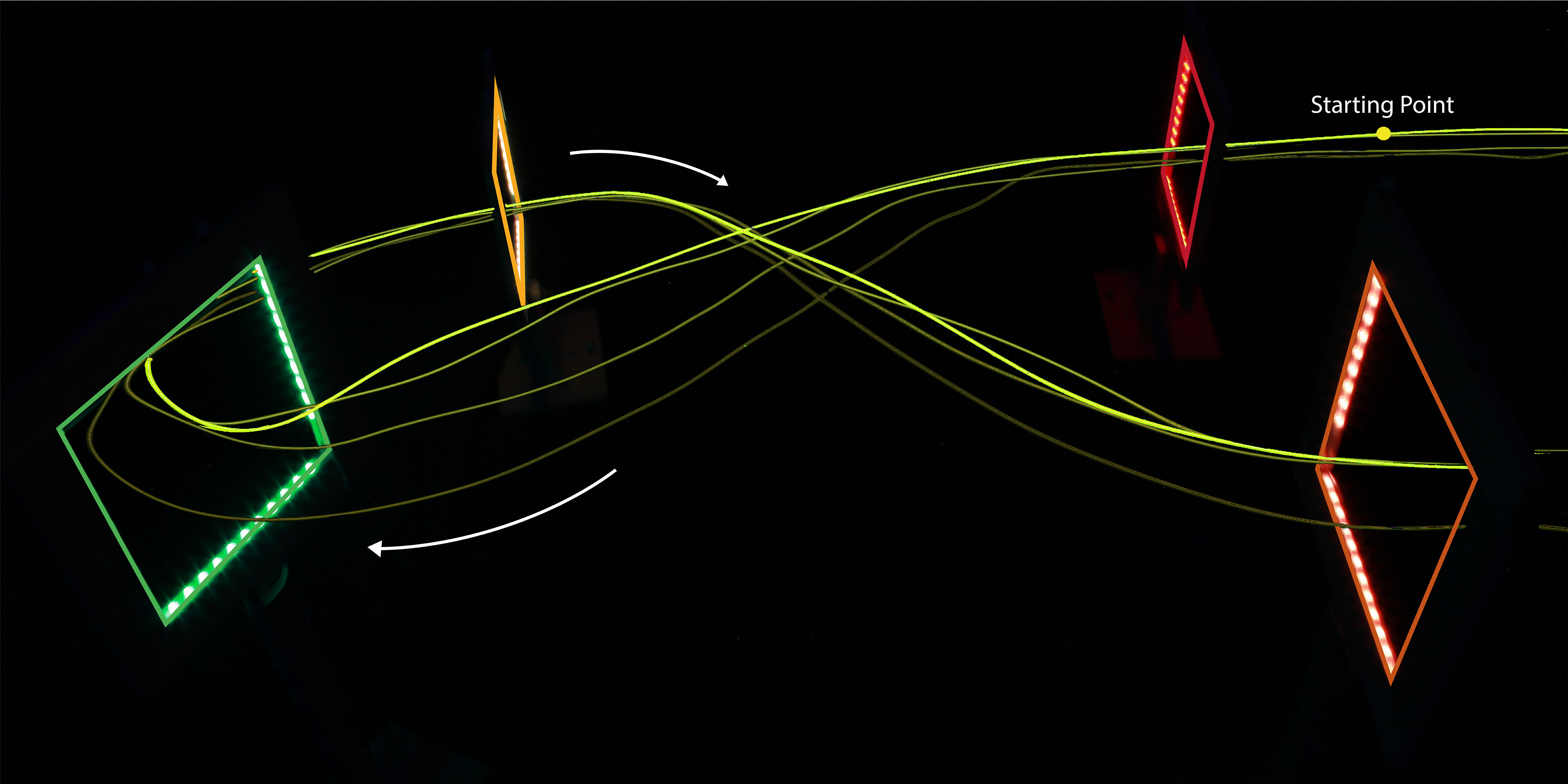}
  \caption{
  LMPC trajectories from initialization (faintest line) to iterations 1, 2, and 3 (brightest). As iterations progress, the path tightens around the gates, and the lap time decreases.
  A video demonstrating the controllers' performance can be found here: \href{http://tiny.cc/lmpc-drone-racing}{http://tiny.cc/lmpc-drone-racing}.} 
  \label{fig:long_exposure}
\end{figure}

Moreover, extensive simulation and real-world experiments further validate the effectiveness of our proposed approach, as shown in~\autoref{fig:long_exposure}, demonstrating that LMPC can refine any initial trajectory while preserving its performance benefits, including those from state-of-the-art controllers like MPCC++.

\section{Related Works}
\label{sec:related-works}
\subsubsection{Drone Racing}
MPC, as well as learning-based methods, have been widely explored in the context of drone racing.
In the domain of model-based control for agile quadrotor flight, nonlinear MPC~(NMPC) has been shown to outperform linear and flatness-based methods~\cite{kamel2017linear, sun2022comparative}, primarily due to the highly nonlinear dynamics of quadrotors. A notable advancement in NMPC is MPCC. Unlike conventional trajectory tracking approaches that rely on predefined timestamps, MPCC optimizes the trade-off between progress along the reference path and contouring error, enabling near time-optimal performance in drone racing~\cite{romero2022model}. The introduction of corridor constraints and a terminal set in MPCC++~\cite{krinner2024mpcc++} further enhances safety and time-optimality by ensuring recursive feasibility in aggressive maneuvers.

While model-based approaches demonstrated strong performance, learning-based methods also emerged as a promising alternative. RL has been successfully applied to high-speed drone racing, with approaches based on PPO achieving competitive performance relative to MPCC++~\cite{song2023reaching, krinner2024mpcc++, kaufmann2023champion, song2021autonomous}. Despite this advancement, learning-based methods present several limitations. They are often sample inefficient, and sim-to-real transfer is not guaranteed. Moreover, generalization to previously unseen tracks remains challenging, often necessitating retraining. Even on familiar tracks, safety guarantees are difficult to establish. Crucially, none of the aforementioned approaches enable safe, high-speed quadrotor flight while incorporating iterative learning mechanisms across trials. This limitation motivates the exploration of LMPC methods, which aim to bridge the gap between iterative learning and the safety guarantees of MPC.

\subsubsection{Iterative Learning Model Predictive Control (LMPC)}
In MPC, a safe set ensures recursive feasibility by defining a region where a feasible control policy keeps the system within constraints indefinitely. While this is crucial for safety in high-speed flight, computing safe sets in practice is challenging in general~\cite{BLANCHINI19991747}. LMPC addresses this challenge by leveraging historical trajectories to construct a safe set, allowing previously successful maneuvers to be reused and refined \cite{rosolia2017learning}. Initially applied to autonomous car racing, this iterative approach has demonstrated consistent performance improvements over multiple laps while requiring significantly fewer training laps compared to RL methods~\cite{rosolia2017autonomous, brunner2017repetitive}. However, existing implementations typically operate at low control frequencies~($\SI{10}{\hertz}$) and rely on Frenet-frame dynamics to constrain the states within the racing zone~\cite{werling2010optimal}.

The application of LMPC to drone racing remains challenging for several reasons. Compared to ground vehicles, quadrotors require significantly higher control frequencies and are subject to more complex nonlinear dynamics, increasing the difficulty of the optimization problem. The authors in~\cite{li2022learning} introduced LMPC for drone racing using the Acados optimization software package~\cite{verschueren2022acados}, achieving a substantial improvement in control frequency compared to previous approaches~\cite{moon2019challenges}. Their work was the first to successfully deploy LMPC on a real quadrotor. However, the method was demonstrated on a relatively simple L-shaped tracking task and did not consistently show performance improvement over iterations, underscoring the difficulty of transferring LMPC’s theoretical guarantees to real-world applications. Recently, the authors in~\cite{calogero2023learning} proposed a local interpolation method for the sampled value function to reduce computational time. While this approach improved efficiency, it relied on a simplified race track representation and was validated only in simulation.

These prior works highlight the potential and challenges of applying LMPC to drone racing. This paper seeks to advance LMPC for drone racing by addressing key shortcomings, including balancing gate traversal with aggressive maneuvers, developing a generalized representation of the racing zone, and ensuring real-time feasibility.

\section{Problem Definition}
\label{sec:problem_def}
We define our drone racing problem in the following. 
The drone behaves as a nonlinear control-affine system:
\begin{equation}
\label{eq:control-affine-sys}
    \dot{\vec{x}}(t) = \vec{f}(\vec{x}(t)) + \vec{G}(\vec{x}(t)) \vec{u}(t) \,,
\end{equation}
where $t \geq 0$ is the time, $\vec{x} \in \R^{9}$ is the state vector, $\vec{u} \in \R^4$ is the control input, and $\vec{f}$ and $\vec{G}$ are Lipschitz continuous functions. The system is subject to nonlinear state and affine input constraints $\vec{x} \in \set{X}$ and $\vec{u} \in \set{U}$, respectively.

From a fixed starting pose $\vec{T}_{is} = \{\vec{p}^{si}_i, \vec{R}_{is}\} \in \SE$~(where~$i$ indicates the inertial frame $I$ and $s$ indicates the starting frame $S$), the drone is tasked to pass through a series of $N_g$ gates. Each gate is defined by a given convex and compact set $\set{T}_n \subset \R^3$ such that $\vec{t}^\intercal \vec{e}_z = 0\,, \forall \vec{t} \in \set{T}_n$ with $\vec{e}_z = \left[0\quad 0\quad 1 \right]^\intercal$ indicating the opening of the gate and a known pose in the inertial frame $\vec{T}_{it_{n}} \in \SE$. Each gate's $z$-direction indicates the direction of traversal, and the drone must traverse all gates in the given order $n \in \{1, \dots, N_g\}$.  We refer to this setup as the track. The task is considered to be executed \emph{successfully} if no constraints are violated~(e.g., no collisions with the gates or the environment). Since we consider the racing setting, the goal is to minimize the time required for the drone to complete the track successfully. The traversal time is the time $t_{N_g}$ indicating when the drone passes through the last gate with index~$N_g$ after starting from the initial pose at time $t_0 = 0$. Mathematically, this can be formulated as a minimum-time optimal control problem:
\begin{align}
\label{eq:time-optimal-control-problem}
  t^*_{N_g}(\vec{p}_s) = \quad &\min_{\vec{u}(t)} \quad t_{N_g} \notag\\
\quad \text{s.t.} \quad & \dot{\vec{x}}(t) = \vec{f}(\vec{x}(t)) + \vec{G}(\vec{x}(t))\vec{u}(t), \quad \notag \\
\quad & \left[x(t_n),~y(t_n),~z(t_n)\right]^\intercal \in \vec{R}_{it_{n}} \set{T}_n + \vec{p}^{t_{n}i}_i \notag \\
\quad & \left[\dot{x}(t_n),~\dot{y}(t_n),~\dot{z}(t_n)\right]^\intercal (\vec{R}_{it_{n}} \vec{e}_z) > 0 \\
\quad & 0 \leq t_1,\, t_n \leq t_{n + 1},\, \forall n \in \N_{1, N_g -1}\notag\\
 \quad & \vec{x}(t) \in \set{X},\, \vec{u}(t) \in \set{U}\,, \forall t \in \left[0, t_{N_g} \right]\notag\\
 \quad & \vec{x}_{\text{pose}}(0) = \vec{T}_{is} \,,\, \dot{\vec{x}}(0) = \vec{0} \notag
 \,,
\end{align}
where $\vec{x}_{\text{pose}} \in \SE$ only contains the part of the drone's state related to its pose, and $\N_{a, b}$ is the shorthand notation for the set of consecutive integers from $a$ to $b$ with $a < b$. 
Since the general minimum-time optimal control problem is challenging to solve, we relax the problem and assume that we are given a suboptimal but successful demonstration of a state and input trajectory that provides a starting point for an improved feasible solution. 

\section{PRELIMINARIES}
\label{sec:background}
In this section, we introduce the drone dynamics and mathematical background for the MPC and LMPC algorithms.

\subsection{Drone Dynamics}
\label{sec:quad-dyn}
The drone's state is defined as $\vec{x} = [\vec{p}_i^\intercal, \vec{v}_i^\intercal, \boldsymbol{\phi}^{ib\intercal}_i]^\intercal \in \mathbb{R}^{9}$, where $\vec{p}_i \in \mathbb{R}^{3}$ is the position of the drone's center of gravity~(CoG) given in the inertial frame $I$, $\vec{v}_i \in \mathbb{R}^{3}$ is the velocity, $\boldsymbol{\phi}^{ib}_i \in \SO$ are the Euler angles that describe the rotation from the body frame $B$ to the inertal frame. 
The control input is the collective thrust-attitude interface $\vec{u} = [f_{\Sigma}, \boldsymbol{\phi}_{\cmd}^\intercal]^\intercal \in \mathbb{R}^{4}$, where $f_{\Sigma}=\sum_{i=1}^4f_i \in \set{U}_f$ is the collective thrust from the four propellers, $\set{U}_f=\left[0, f_{\Sigma, \max} \right]$, and $\boldsymbol{\phi}_{\cmd} \in \SO$ are the commanded Euler angles. Therefore, $\set{U} = \set{U}_f \times \SO$. For brevity, we drop the frame indices as all the variables are expressed in the inertial frame, which leads to the following state and input vector expressions: $\vec{x} = [x, y, z, v_x, v_y, v_z, \phi, \theta, \psi]^\intercal\,,\text{ and}~ \vec{u} = [f_\Sigma, \phi_{\cmd}, \theta_{\cmd}, \psi_{\cmd}]^\intercal \,.$ Then, the translational dynamics can be derived from Newton’s law, and the rotational dynamics can be modeled as three independent first-order integrators:
\begin{equation}
    \dot{\vec{x}} = \vec{f}(\vec{x}) + \begin{bmatrix}
        \vec{0}_{3 \times 1} & \vec{0}_{3 \times 3} \\
        \vec{G}_1(\vec{x}) & \vec{0}_{3 \times 3} \\
        \vec{0}_{3 \times 1} & \vec{G}_{2}
    \end{bmatrix} \begin{bmatrix}
        f_{\Sigma} \\ \boldsymbol{\phi}_{\cmd}
    \end{bmatrix}\,,
\end{equation}
where we dropped the dependency on $t$, $\vec{f}(\vec{x}) = \begin{bmatrix}
    v_x & v_y & v_z & 0 & 0 & -g & \alpha_{\phi} \phi & \alpha_{\theta} \theta & \alpha_{\psi} \psi
\end{bmatrix}^\intercal$ and  
\begin{equation*} \label{sys_dynamics}
  \vec{G}_1(\vec{x}) = \frac{1}{m}
    \begin{bmatrix} 
    \co\phi\,\s\theta\,\co\psi + \s\phi\, \s\psi\\ 
    \co\phi\,\s\theta\, \s\psi - \s\phi\, \co\psi\\ 
    \co\phi\,\co\theta 
    \end{bmatrix} \,,~
  \vec{G}_{2} = 
    \text{diag}(\beta_{\phi}\,, \beta_{\theta}\,,\beta_{\psi} )\,,
\end{equation*}
where $\s$ and $\co$ are shorthand notations for $\sin$ and $\cos$, respectively, $m$ is the drone's mass, and the coefficients in the single-integrator model are $[\alpha_{\phi}, \alpha_{\theta}, \alpha_{\psi}, \beta_{\phi}, \beta_{\theta}, \beta_{\psi}]$. The parameters are identified from real-world experiments using MATLAB's System Identification package, see~\autoref{sec:eval-section}.

\subsection{MPC and Iterative Learning MPC}
MPC is a model-based control strategy that leverages the system's model to predict the effect of the chosen control actions on the system's state over a finite horizon $N > 0$ of discrete time steps~\cite{Borrelli2017}. At every time step, MPC solves the following constrained optimization problem:
\begin{subequations} \label{eq:MPC-basic}
\begin{align}
J_{k}^{*}(\vec{x}_k) =& \min_{\{\vec{u}_{k+i|k}\}} \sum_{i=0}^{N-1} h(\vec{x}_{k+i|k}, \vec{u}_{k+i|k}) + V_f(\vec{x}_{N|k}) \label{eq:MPC_cost}\\
\text{s.t.} \quad
&\vec{x}_{k+i+1|k} = \vec{f}(\vec{x}_{k+i|k}) + \vec{G}(\vec{x}_{k+i|k}) \vec{u}_{k+i|k}, \label{eq:mpc-system_dynamics}\\
&\vec{x}_{k+i|k} \in \set{X}\,,\vec{u}_{k+i|k} \in \set{U}\,,\vec{x}_{k|k} = \vec{x}_k\,, i \in \set{N}_{0, N-1} \label{eq:mpc-state_input_constraints}\\
&\vec{x}_{k+N|k} \in \set{X}_f \label{eq:MPC_terminal_constraint} \,,
\end{align}
\end{subequations}
where~\eqref{eq:mpc-system_dynamics} is now a discrete-time version of the dynamics in~\eqref{eq:control-affine-sys} with sampling time $\frac{1}{f_d}$, $h$ is the stage cost, and $V_f$ and $\set{X}_f$ are the terminal cost function and the terminal constraint set, respectively, which are often referred to as the terminal ingredients. The terminal ingredients' purpose is to account for the MPC's short-sightedness and certify that the resulting control law is stabilizing~($V_f$ is a control Lyapunov function) and feasible at every time step~($\set{X}_f$ is a control invariant set, i.e., the system can be kept inside $\set{X}_f$ for all time $t > 0$ if $\vec{x}(0) \in \set{X}_f$)~\cite{Borrelli2017}. Only the input sequence's first control input is applied to the system, and the problem in~\eqref{eq:MPC-basic} is solved again at the next time step in a receding horizon fashion. 

For general nonlinear control affine systems with constraints, determining the terminal ingredients is a challenging problem~\cite{BLANCHINI19991747}. Therefore, MPCs are often used without terminal ingredients~\cite{QIN2003733}. LMPC alleviates this problem by constructing the terminal ingredients from previously successful iterations~\cite{rosolia2017learning}. LMPC achieves non-decreasing improvement with every additional iteration and is therefore well-suited for minimum-time problems. Next, we detail how LMPC generates $V_f$ and $\set{X}_f$ in~\eqref{eq:MPC-basic} from data. 

LMPC is an iterative approach that uses previous state and input trajectories~\cite{rosolia2017learning}. The trajectories are given as discrete sets
  $\vec{X}^j = \{\vec{x}_0^j, \vec{x}_1^j, \ldots, \vec{x}_{T_j}^j\} $ and 
  $\vec{U}^j = \{\vec{u}_0^j, \vec{u}_1^j, \ldots, \vec{u}_{T_{j-1}}^j\}\,,$
respectively,  where $j$ denotes the iteration and $T_j$ denotes the index at which the task is completed.
At the $j$-th iteration, a sampled safe set $\safeset^j$ is constructed as the union of states from all successful previous iterations $G^j$ as
\begin{equation*}
  SS^j = \bigcup_{i \in G^j}\vec{X}^i = \bigcup_{i \in G^j} \bigcup_{k=0}^{T_i} {\vec{x}_k^i} \,.
\end{equation*}
As shown in~\cite{rosolia2017learning}, $\safeset^j$ is a control invariant set, as from any state in $\safeset^j$ there exists a feasible sequence of control inputs that achieves the task. 
Based on the past iterations, the cost-to-go for state $\vec{x}_k^j$ in iteration $j$ at step $k$ is
\begin{equation*}
  J_{\N_{k,T_j}}^j(\vec{x}_k^j) = \sum_{t=k}^{T_j} h(\vec{x}_t^j, \vec{u}_t^j) \,,
\end{equation*}
which is the cumulative cost over all future states when following trajectory $\vec{X}^j$ from state $\vec{x}_k^j$ onwards.
The optimal cost-to-go $Q^j(\vec{x})$ of a state $\vec{x} \in \safeset^j$ at iteration $j$ is 
\begin{subequations}
\begin{align}
  Q^j(\vec{x})& = \begin{cases}
  \min_{(i,k) \in F_j(\vec{x})} J_{\N_{k,T_i}}^i(\vec{x}), & \text{if } \vec{x} \in SS^j \\
  +\infty, & \text{if } \vec{x} \notin SS^j \end{cases}\,,\\ 
  F^j(\vec{x}) &= \left\{ (i, t) \,:\, i \in G^j\,, t \in \N_{0, T_i}\,, \vec{x} = \vec{x}_t^i\,, \, \vec{x}_t^i \in SS^j \right\} \,.
\end{align}
\end{subequations}
In~\cite{rosolia2017learning}, it is proven that $Q^j$ is a control Lyapunov function, which assigns the minimum cost-to-go to each state in $\safeset^j$.

\subsection{Safe Set Relaxations}
While $\safeset^j$ and $Q^j$ could already be substituted into~\eqref{eq:MPC-basic}, this would lead to a mixed-integer programming problem, and the sampled safe set can grow unboundedly. Typically, a~local convex safe set is constructed to reduce the computational complexity. As proposed in~\cite{rosolia2019learning}, we select the indices of the $k$-nearest neighbours $\mathcal{K}^{j}(\bar{\vec{x}})$ of $\bar{\vec{x}}$ in $\safeset^j$, where $k > 0$ and $\bar{\vec{x}} = \vec{f}(\vec{x}_{k+N|k-1}, \vec{u}_{k+N-1|k-1}) \approx \vec{x}_{k+N|k}$ is an estimate of the terminal state at the current time step by making a one-step prediction based on the previous time step. 

The local safe set $\mathcal{LS}^j$ and local optimal approximation of the cost-to-go $Q_l^j$ are defined as
\begin{subequations}
\begin{align}
  &\mathcal{LS}^{j}(\bar{\vec{x}}) = \bigcup_{(i, k) \in \mathcal{K}^{j}(\bar{\vec{x}})}\vec{x}_k^i \in SS^{j}, \label{def:local_asfety_set}\\ 
  &Q^j_l(\vec{x}, \bar{\vec{x}}) =
  \begin{cases}
  \min\limits_{\vec{x} \in \mathcal{LS}^j(\bar{\vec{x}})} J_{\N_{k,T_i}}(\vec{x}), & \text{if } \vec{x} \in \mathcal{LS}^j(\bar{\vec{x}}) \\
  +\infty, & \text{if } \vec{x} \notin \mathcal{LS}^j(\bar{\vec{x}})
  \end{cases} \,,
\end{align}
\end{subequations}
which are subsequently turned into convex approximations $\mathcal{CS}$ and $\tilde{Q}_l$, respectively, as
\begin{subequations}
\begin{align}
\mathcal{CS}^j &= \text{conv}(\mathcal{LS}^j) = \sum_{(i, k) \in \mathcal{K}^j(\bar{\vec{x}})}\vec{x}_{k}^{i}\vec{\lambda}_k^i \,,\\
\tilde{Q}^j_l(\vec{x}) &= \text{conv}(Q^j(\vec{x}))= \min_{\lambda_k^i \geq 0}  \sum_{(i, k) \in \mathcal{K}^j(\bar{\vec{x}})}J_{\N_{k,T_i}}(\vec{x}_k^i) \vec{\lambda}_k^i 
\end{align}
\end{subequations}
where $\vec{\lambda} \in \R^k$ denotes the weight vector for each state in $\mathcal{LS}^j$ and satisfies $\sum_{(i, k) \in \mathcal{K}^j(\bar{\vec{x}})}\vec{\lambda}_k^i = 1$.

Finally, we formulate the following constrained finite-time optimal control problem:
\begin{align}
\label{LMPC_formulation_relaxation}
J_{k}^{*,j}(\vec{x}_k^j) = &\min_{\{\vec{u}_{k+i|k}\}, \vec{\lambda} \geq 0} \sum_{i=0}^{N-1} h(\vec{x}_{k+i|k}, \vec{u}_{k+i|k}) + \tilde{Q}^{j-1}_l(\vec{x}_{k+N|k}) \notag\\
\text{s.t.} \quad
& \eqref{eq:mpc-system_dynamics}, \eqref{eq:mpc-state_input_constraints}\\
&\vec{x}_{k+N|k} = \sum_{(i, k) \in \mathcal{K}^{j-1}(\bar{\vec{x}})}\vec{x}_{k}^{i}\vec{\lambda}_k^i\,, \sum_{(i, k) \in \mathcal{K}^{j-1}(\bar{\vec{x}})}\vec{\lambda}_k^i = 1\,, \notag
\end{align}
where we replaced the terminal constraint set and cost in~\eqref{eq:MPC-basic} with $\mathcal{CS}^j$ and $\tilde{Q}^{j-1}_l$, respectively.


\section{METHODOLOGY}

\begin{figure}[t]
  \centering
  \includegraphics[width=0.9\linewidth]{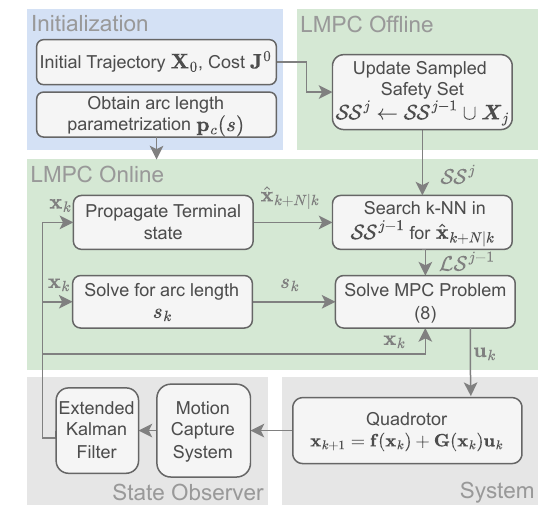}
  \caption{Block diagram of our proposed LMPC framework. The algorithm initializes and updates offline with the collected trajectory $\vec{X}_j$. In the online part, the controller receives the current state $\vec{x}_k$, computes the arc-length $s_k$, estimates the terminal state $\hat{\vec{x}}_{k+N|k}$, and extracts a local safe set. Control action $\vec{u}_k$ is then computed and passed to the drone.}
  \label{fig:block_diagram}
\end{figure}

In this section, we introduce our modifications to LMPC for drone racing, as illustrated in Fig.~\ref{fig:block_diagram}.

\subsection{Race Track Definition and State Constraints}
The race track is defined by a corridor with varying cross-section that passes through all the gates in the specified order. The corridor is designed to satisfy all the state constraints posed in~\eqref{eq:time-optimal-control-problem}, including collisions with any gate or the environment. In this section, we first detail how we generate an arc-length parametrized central path, which we use to define the spatial constraints in the form of a corridor.

The central path $\vec{p}$ is generated based on a piecewise cubic Hermite interpolation~(CHI) through the center of the gates. Each piece of the CHI profile $H^3_n$ satisfies: 
\begin{equation}
\begin{cases}
H^3_n(l_n) = \vec{p}_i^{t_ni}, \quad H^3_n(l_{n+1}) = \vec{p}_i^{t_{n+1}i}
\\ 
\dot{H}^3_n(l_n) = \vec{R}_{it_{n}} \vec{e}_z, \quad \dot{H}^3_n(l_{n+1}) = \vec{R}_{it_{n + 1}} \vec{e}_z
\end{cases}\,, n \in \N_{1, N_g-1},
\label{CHI_constraints}
\end{equation}
where $l$ denotes the interpolation variable. CHI ensures smooth transitions between the given gate positions and that the first derivative is aligned with the $z$ axis of each gate. 

Once the central path $\vec{p}$ in the Cartesian space has been obtained, we parameterize it using an arc length variable, $s \in \R$, as $\vec{p}_{\text{c}}(s)$. The arc length defines the progress on the central path with $s = s_{\min}$ and $s = s_{\max}$ indicating the beginning $\vec{p}_{\text{c}}(s_{\min}) = \vec{p}_i^{si}$ and end of the path $\vec{p}_{\text{c}}(s_{\max}) = \vec{p}_i^{t_{N_{g}}i}$, respectively. 
Based on the arc-length parametrized central path $\vec{p}_c(s)$, we define the corridor safety constraint $\vec{g}_c(\vec{x}) \leq \vec{0}$ as the circular cross-section along the central path:
\begin{equation}
    \mathcal{A}_{\text{c}}(s) = \{\vec{p} \in \R^3 \,:\, \|\vec{p}-\vec{p}_{\text{c}}(s)\| \leq R_{\text{c}}(s)\,,\,  (\vec{p} - \vec{p}_c(s))^\intercal \vec{T}(s)=0\} \,, 
    \label{corridor safety constraint}
\end{equation}
where $R_{\text{c}}$ is the radius of the circular cross-section, and $\vec{T}(s)$ is the tangent vector of the centerline. 
We define the smallest radius at the gate positions as $R_{{\text{c}}, \text{gate}}$ and the largest radius in between the gates as $R_{{\text{c}},\max}$. We smoothly transition between the radius using sigmoid functions as in~\cite{krinner2024mpcc++}.
This allows us to trade off between gate passing and exploration.

To apply the cross-section constraints, we need to keep track of the system's position in terms of the arc length. Therefore, we augment the state vector to contain the arc length $s$ as $\vec{x}_{\text{aug}} = \begin{bmatrix}
    s & \vec{x}^\intercal
\end{bmatrix}^\intercal$
with the dynamics equation $\dot{s} = \vec{v}^\intercal \vec{T}(s)$ with $\vec{T}(s) = \vec{\dot{p}}_c(s) /\|\vec{\dot{p}}_c(s)\|,$ 
where $\vec{v}=[v_x, v_y, v_z]^\intercal$.

This centerline construction method brings an important advantage: Given the gate positions, order, and orientations, a reference trajectory can be automatically generated. This enables strong generalization to arbitrary and unseen track layouts, which will be validated in the experimental section.

Compared to~\cite{rosolia2017autonomous}, we do not propagate the dynamics using a moving frame based on the arc length and only propagate the arc length itself, which we correct using feedback at each time step. Therefore, we can avoid wind-up errors and any potential singularities present in such a representation. 

\subsection{Estimating the Current Arc Length}
To augment the state with the arc length, we must determine the system's arc length on the central path at each time step. 
The authors in~\cite{romero2022model} propose recording the predicted arc length at each time step and using it for initialization in the subsequent step. However, this approach can suffer from wind-up errors, leading to poor performance, especially at lower control frequencies. 
Therefore, we propose an efficient method that determines the current arc length $s_{k}$ based on the current Cartesian position. 
This requires identifying the point on the curve $\vec{p}_{\text{c}}(s)$ that minimizes the Euclidean distance to the current position $\vec{p}_{k}$. We solve this problem in three steps: First, we discretize the arc length interval $[s_{\min}, s_{\max}]$ into equally sized bins $\mathcal{S}=\{s_i\}_{i=1}^{N_s-1}$, 
where $N_s$ denotes the number of bins. Second, assuming that the arc length from the previous time step $s_{k-1}$ is available and the current state is close to the system's state at the previous step, we can search for the index $i^*$ of the nearest neighbor of $s_{k-1}$ in $\mathcal{S}$. 
Finally, we solve the following continuous optimization problem with an L-BFGS-B solver to determine the arc length associated with the minimal Euclidean distance:
\begin{equation}
    s_{k} = \arg\min_s \|\vec{p}_c(s) - \vec{p}_{k}\|, \quad s\in [s_{i^*-1}, s_{i^*+1}], \label{solve_s_current_2}
\end{equation}
\noindent where $s_{i^*-1}$ and $s_{i^*+1}$ denote $s_{i^*}$'s previous and next node.

Note that the second step typically entails the greatest computational overhead. However, this cost can be significantly reduced by storing all of $\mathcal{S}$'s elements offline in a $k$-d tree and performing the queries online.

\subsection{Cost Design}

First, we introduce the time-optimal cost design with a control input penalty $l_t(\vec{u}) = c + \|\vec{u}\|_{\vec{R}}^2$, where $c > 0$ penalizes the use of additional steps, $\vec{R}$ is a positive definite weight matrix, and $\|\vec{u}\|_{\vec{R}}^2 = \vec{u}^\intercal \vec{R} \vec{u}$.

Using the above cost, we observed that the LMPC tends to select control inputs that take shortcuts to improve the track time. Although this is desired, it may reduce the success rate of passing through all gates. Overly aggressive corner-cutting may lead to collisions with the gate boundaries. Therefore, we introduce a penalty term on the lateral deviation:
\begin{equation}
    l_d(\vec{x}) = \bigg\Vert \frac{\vec{p} - \vec{p}_c(s)}{R_c(s)}\bigg\rVert_{\vec{Q}_{d}}^2 , 
    \label{LMPC_DR_stage_cost_lateral_deviation}
\end{equation}
where $\vec{Q}_{d}$ is a positive definite weight matrix. For additional flexibility, we introduce a lateral deviation penalty $\gamma$ as a function of the arc length. This penalty is larger when approaching each gate to encourage the drone to be closer to the central path. We construct this penalty using two mirrored sigmoid functions $\sigma_{\text{in}}$ and $\sigma_{\text{out}}$ for each gate as $\gamma(s) = \sum_{n=1}^{N_g} \gamma_n\sigma_{\text{in}}^n(s) \sigma_{\text{out}}^n(s)$, with $\sigma_{d}^n(s) = \text{sigmoid}(\rho_{d}^n(s)) \,,$ where $\gamma_n$ denotes the weight coefficient for each gate, 
$\rho_{d}^n(s) = k_{d}(s - s_{g,n})+6$, for $d \in \{\text{in}, \text{out} \}$, 
$k_{\text{in}} > 0$ and $k_{\text{out}}<0$ are slope factors, and $s_{g,n}$ is the arc length at each gate. 

Combined, this enables our LMPC to trade off between gate passing and minimum lap time:
\begin{equation}
    h(\vec{x}, \vec{u}) = l_t(\vec{u}) + \gamma(s)l_d(\vec{x}) \,.
    \label{LMPC_DR_stage_cost_combined}
\end{equation}

\subsection{Modified Local Safe Set}
In the experiments, we found that adding the lateral deviation cost was insufficient to prevent the drone from cutting corners completely. This issue arises because LMPC always selects terminal states from the convex safe set $\mathcal{CS}^{j-1}(\bar{\vec{x}})$. 
Given our cost design, LMPC consistently favors states that cut corners, i.e., farther away from the central path. Consequently, the states in the safe set tend to spatially concentrate in a small region. 
Although the cost function encourages minimizing the deviation from the centerline, there are typically no states in the safe set that allow the LMPC to converge back to the centerline from the current position, rendering the penalty term ineffective.

\begin{figure}[t]
  \centering
  \includegraphics[width=\columnwidth, clip]{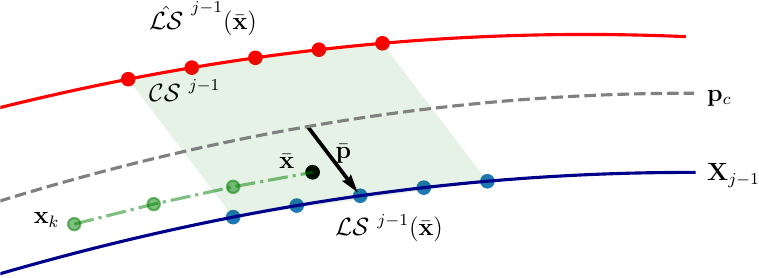}
  \caption{ At the current state $\mathbf{x}_k$, the modified sampled safe set is determined. First, the terminal state $\bar{\mathbf{x}}$ is estimated and the previous trajectory $\mathbf{X}^{j-1}$ is translated beyond the center line $\mathbf{p}_c$. The $K$-nearest neighbors~($K$NN) are extracted on both the shifted and non-shifted trajectories. These states are then used to construct the augmented convex safe set $\mathcal{CS}^{j-1}$.}
  \label{fig:safe_set}
\end{figure}

To address this problem,~\cite{Lukas2018} leverages multiple varied initializations to generate a more diverse sampled safe set. Instead, we find that a single demonstration is sufficient if we artificially introduce states on the opposite side of the central path. We achieve this by shifting all the states in the current local safety set beyond the central path, as shown in~\autoref{fig:safe_set}. Given the current local safety set $\mathcal{LS}^{j-1}(\bar{\vec{x}})$, we can compute the average distance from each position in the local safety set to the closest point on the centerline:
\begin{equation*}
    \overline{\vec{p}} = \frac{1}{K} \sum_{k=1}^{K}(\vec{p}_k - \vec{p}_c(s_k)), \quad \bar{s} = \frac{1}{K} \sum_{k=1}^K s_k\,, \quad [\vec{p}_k, s_k] \in \mathcal{LS}^{j-1}(\bar{\vec{x}}),
    \label{lss_modeification:average distance array}
\end{equation*}

\noindent where $K = |\mathcal{LS}^{j-1}(\bar{\vec{x}})|$ is the cardinality of the local safe set. Then, the opposite side of the central path is:
\begin{equation}
    \mathcal{A}_o(\bar{s}) = \{ \vec{p} \in \R^3\,:\, {}\vec{p} \in \mathcal{A}_c(\bar{s}) \,,\, (\vec{p} - \vec{p}_c(\bar{s}))^\intercal \overline{\vec{p}} \leq 0 \}\,.
    \label{lss_modeification:opposive area}
\end{equation}

Translating the local safe set to any position opposite the central path at $\bar{s}$ yields $\hat{\mathcal{LS}}^{j-1}$. To not prioritize these states as terminal states, we over-approximate their cost-to-go by adding a term that grows quadratically with the Euclidean distance from the original state:
\begin{align}
  \hat{J}_{k,T^j} = J_{k,T^j} + \|\hat{\vec{p}}_k - \vec{p}_k\|_{\vec{K}}^2,~
   \vec{p}_k \in \mathcal{LS}^{j-1}(\bar{\vec{x}}),~\hat{\vec{p}}_k \in \hat{\mathcal{LS}}^{j-1}(\bar{\vec{x}})\,, \notag
\end{align}
where $\vec{K}$ is a positive definite weight matrix.

\section{EVALUATIONS}
\label{sec:eval-section}

In this work, we utilize the Crazyflie 2.1 quadrotor platform for all real-world experiments. Our method runs on an offboard desktop computer equipped with an Intel i7-11700H CPU. The drone's and gates' poses are tracked using a motion capture system at $\SI{200}{\hertz}$. An extended Kalman filter~(EKF) fuses motion capture data with the drone's dynamic model. System identification from real-world data yields the model parameters $[\alpha_{\phi}, \alpha_{\theta}, \alpha_{\psi}] = [-6.00, -3.96, 0.0]$ and
$[\beta_{\phi}, \beta_{\theta}, \beta_{\psi}] = [6.21, 4.08, 0.0]$. The yaw angle $\psi$ is held constant during all experiments to simplify control. 

\subsection{Implementation Details}

The proposed LMPC controller is implemented using the acados framework (v0.4.1)~\cite{verschueren2022acados} and runs at a control frequency of $\SI{30}{\hertz}$, whereas the baseline PID and MPCC++ controllers operate at a higher frequency of $\SI{90}{\hertz}$. We configure acados with the SQP method as the nonlinear programming solver and use the full condensing HPIPM backend for the QP subproblem. The solver mode is set to \texttt{BALANCE}, which provides a trade-off between speed and robustness. To further enhance the real-time performance of LMPC, we limit the maximum number of SQP and QP iterations to 5 and 20, respectively, and set the convergence tolerance to $10^{-4}$.

\subsection{Solver Runtime Statistics}

To ensure real-time feasibility of our proposed LMPC variant, we analyze its computational cost under different hyperparameter configurations. Compared to the vanilla LMPC, our proposed variant introduces an additional runtime module for arc-length estimation. For a global safe set consisting of 631 states, directly computing $s$ via brute-force linear search takes on average $4.21 \pm 0.28$ ms per call. To reduce this overhead, we employ a $k$-d tree structure, which reduces the estimation to $0.68 \pm 0.17$ ms and is negligible compared to the overall solver time. 
Therefore, we always use the $k$-d tree in our implementations. 
Table~\ref{tab:solver_timing} reports the average runtime of the proposed LMPC variant as a function of the prediction horizon $N$ and the size of the safe set $K$. Based on this trade-off between computation time and performance, we select $(N, K) = (8, 20)$ as the default setting for the following simulation and real-world experiments. This setting offers a good balance between long-term prediction and optimization feasibility, while maintaining a control frequency of 30 Hz.

\begin{table}[t]
\caption{Solver runtime statistics for our proposed LMPC variant under varying prediction horizons $N$ and safe set sizes $K$. Each entry reports mean and standard deviation in $\si{\milli \second}$.}
\label{tab:solver_timing}
\centering
\begin{tabular}{c|c|c|c}
\toprule
\textbf{Hyperparam.} & \textbf{$K{=}10$} & \textbf{$K{=}15$} & \textbf{$K{=}20$} \\
\midrule
\textbf{$N{=}5$} & $12.13 \pm 1.42$ & $13.95 \pm 1.69$ & $16.66 \pm 2.28$ \\
\midrule
\textbf{$N{=}10$} & $20.41 \pm 3.24$ & $23.61 \pm 2.45$ & $34.26 \pm 5.18$ \\
\midrule
\textbf{$N{=}15$} & $37.21 \pm 6.44$ & $55.92 \pm 6.75$ & $72.24 \pm 7.31$ \\
\bottomrule
\end{tabular}
\end{table}

\subsection{Ablation Studies in Simulation}

In this section, we conduct ablation studies on both the original LMPC algorithm and our proposed modifications in simulation. The drone racing environment used in the simulations follows the gate configuration in the Split-S track from~\cite{song2023reaching} scaled down by a factor of four.

\subsubsection{Ablation Study on LMPC Hyperparameters}

We study the impact of key hyperparameters in the original LMPC formulation~\eqref{LMPC_formulation_relaxation}, focusing on the discretization frequency $f_d$ and the size of the safe set $K$. We initialize our LMPC with a PID controller tracking the center line at a constant velocity $\rVert \vec{v}\lVert=\SI{0.5}{\frac{\meter}{\second}}$, and then perform LMPC for eight iterations and analyze the trajectory at the final lap. The corridor safety constraint is disregarded in this study to isolate the effect of the algorithm’s hyperparameters. We fix $N=8$ and conduct the experiments with $f_d \in \{\SI{16}{\hertz}, \SI{20}{\hertz}, \SI{24}{\hertz}\}$ and $K\in \{10, 15, 20\}$. 
\begin{figure}[t]
  \centering
  \includegraphics[width=\linewidth]{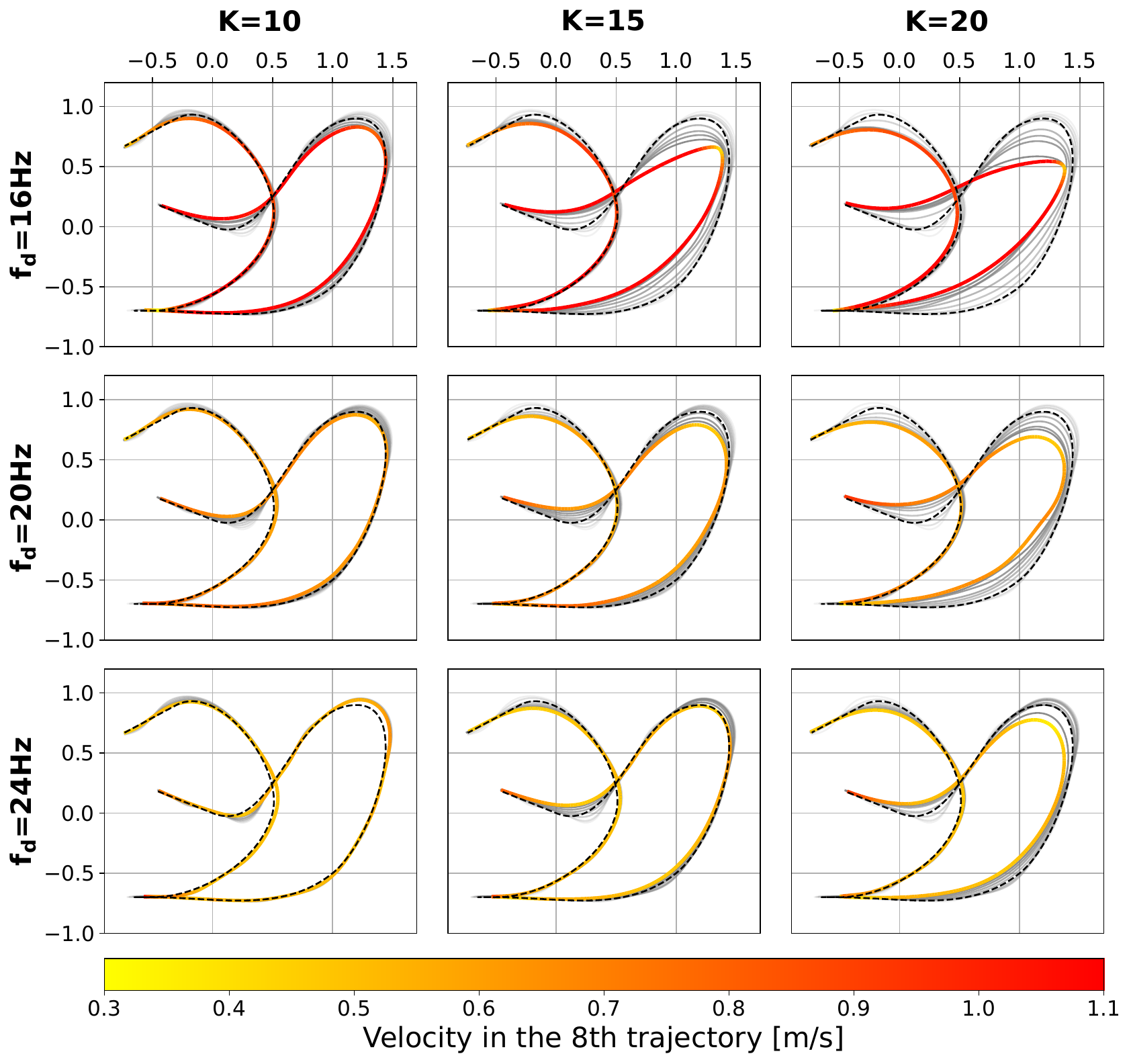}
  \caption{Ablation study on LMPC hyperparameters. The experiments vary the discretization frequency $f_d$ and the safe set size $K$. The dashed black line represents the centerline, while the solid gray lines show the trajectories over iterations. The trajectory in the last iteration is colored to represent 3D velocity, with warmer colors indicating higher velocities.}
  \label{Ablation Study on LMPC Hyperparameter}
\end{figure}
The results~(see~\autoref{Ablation Study on LMPC Hyperparameter}) indicate that LMPC iterations primarily lead to two effects: acceleration and shortcutting. 
The acceleration effect is primarily driven by a smaller discretization frequency $f_d$, while the influence of $K$ is relatively minor. In contrast, shortcutting behavior is promoted by both a smaller $f_d$ and a larger safe set size $K$. Unlike acceleration, which is purely beneficial, shortcutting can be detrimental in drone racing.
Due to the existence of model mismatch and state estimation errors, overly aggressive shortcuts increase the risk of collisions. Therefore, in the context of drone racing, choosing relatively small $f_d$ and $K$ with an appropriate $N$ yields better overall performance and safety.

\begin{figure*}[t]
\centering
\includegraphics[width=\linewidth]{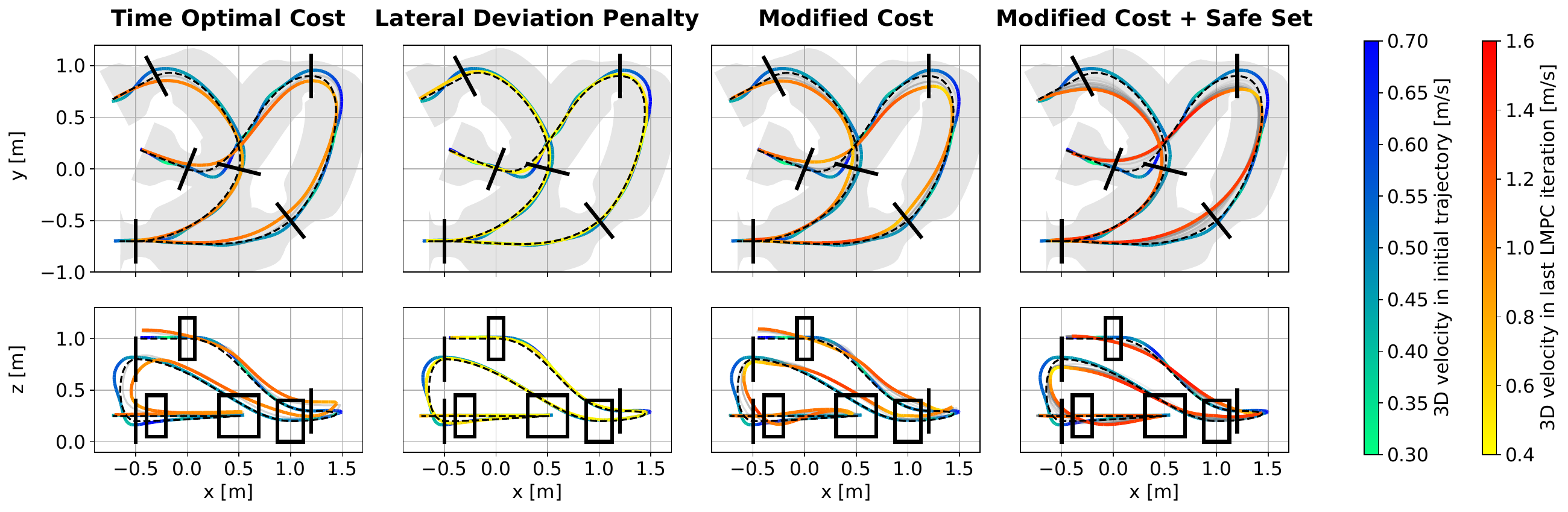}
\caption{Simulated racing performance for LMPC with the different components we proposed. The initial and final trajectories are colored using the left and right color bars, respectively. The central path and the corridor racing zone are represented by a dashed black line and the shaded zone in gray, respectively.}
\label{Ablation study on cost modifications}
\end{figure*}

\begin{table*}[t]
\caption{Comparing the cost and safe set modifications using the lap time $t^j_{N_g}$ and the fraction of successfully passed gates $N_{passed}/N_g$.}
\label{tab:Ablation study on cost modifications}
\centering
\begin{tabular}{c||cc|cc|cc|cc}
\toprule
\multirow{3}{*}{\textbf{Lap}} 
 & \multicolumn{2}{c}{\textbf{Only Time Optimal Cost}} 
 & \multicolumn{2}{c}{\textbf{Only Lateral Deviation Penalty}} 
 & \multicolumn{2}{c}{\textbf{Modified Cost}} 
 & \multicolumn{2}{c}{\textbf{Modified Cost + Safe Set}} \\ 
 & {Lap Time} & Success Rate 
 & {Lap Time} & Success Rate 
 & {Lap Time} & Success Rate 
 & {Lap Time} & Success Rate \\ 
 & {$t^j_{N_g}$ [\si{\second}]} & $N_{passed}/N_g$
 & {$t^j_{N_g}$ [\si{\second}]} & $N_{passed}/N_g$
 & {$t^j_{N_g}$ [\si{\second}]} & $N_{passed}/N_g$
 & {$t^j_{N_g}$ [\si{\second}]} & $N_{passed}/N_g$\\ 
\midrule
0 & 23.24 & 7/7 (all passed) & 23.24 & 7/7 (all passed) & 23.24 & 7/7 (all passed) & 23.24 & 7/7 (all passed)\\
1 & 12.19 & 7/7 (all passed) & 23.05 & 7/7 (all passed) & 14.05 & 7/7 (all passed) & 13.44 & 7/7 (all passed)\\
2 & \textbf{10.04} & \textbf{7/7 (all passed)} & \textbf{22.44} & \textbf{7/7 (all passed)} & 11.98 & 7/7 (all passed) & 11.75 & 7/7 (all passed)\\
3 & \textcolor{gray}{~7.31} & \textcolor{gray}{5/7 (gates 4\&5)} & 22.87 & 7/7 (all passed) & 10.53 & 7/7 (all passed) & 10.65 & 7/7 (all passed)\\
4 & {-} & {-} & 23.49 & 7/7 (all passed) & ~\textbf{9.64} & \textbf{7/7 (all passed)} & ~9.92 & 7/7 (all passed)\\
5 & {-} & {-} & 23.11 & 7/7 (all passed) & ~\textcolor{gray}{9.03} & \textcolor{gray}{6/7 (gate 5)~~~~~} & ~9.39 & 7/7 (all passed)\\ 
$\vdots$ & $\vdots$ & $\vdots$ & $\vdots$ & $\vdots$ & $\vdots$ & $\vdots$ & $\vdots$ & $\vdots$\\
9 & {-} & {-} & 24.02 & 7/7 (all passed) & {-} & {-} & ~8.58 & 7/7 (all passed)\\
10 & {-} & {-} & 22.96 & 7/7 (all passed) & {-} & {-} & ~\textbf{8.47} & \textbf{7/7 (all passed)}\\

\bottomrule
\end{tabular}
\label{Statistics of cost ablation}
\end{table*}

\subsubsection{Ablation study on cost and safe set modifications}
While a small value for $K$ may be desired in terms of performance, it also further constrains the optimization problem, which can lead to infeasibility. Therefore, we typically have to choose a larger value for $K$ in real-world experiments. 
We proposed modifications to alleviate this issue. In this part, we conduct an ablation study to validate their effectiveness. We use four groups based on the same PID initialization as before, each incorporating different components of our modified LMPC algorithm to analyze their impact on the iterative performance. To highlight the effect of the introduced modifications, we set $K = 20$. For each setting, LMPC iterates until at least one gate is missed, the drone collides with a gate, or ten iterations are reached. The final trajectory from each group is used for comparative analysis, with the results shown in \autoref{Ablation study on cost modifications} and \autoref{tab:Ablation study on cost modifications}. The results indicate that the fastest lap time reduction occurs with the time-optimal cost alone, but the drone starts missing gates after only two iterations. Penalizing only lateral deviation keeps the trajectory near the centerline but shows little improvement. Introducing adaptive lateral deviation penalty slows initial progress yet achieves a shorter lap time of \SI{9.64}{\second}, albeit with gate misses after five iterations. Only by combining the modified cost and local safe set does LMPC converge reliably, reaching the shortest lap time of \SI{8.47}{\second}.

\subsection{Experimental Results}
\label{sec:real-world-exps}

In this section, we present the results of iterative improvement from LMPC with initial trajectories generated by different controllers and hyperparameter combinations. To demonstrate the broad applicability of our approach and highlight the impact of different initial trajectories on performance, we select a PID controller tracking the centerline with $\lVert \vec{v}\rVert=\SI{0.5}{\frac{\meter}{\second}}$, as well as the state-of-the-art MPCC++ controller with three tuning settings ranging from conservative $\mu=0.02$ to aggressive $\mu=0.1$. Note that parameter $\mu$ in the MPCC++ is a weight on the expected progress factor along the trajectory. 

\begin{figure}[t]
\centering
\includegraphics[width=\linewidth]{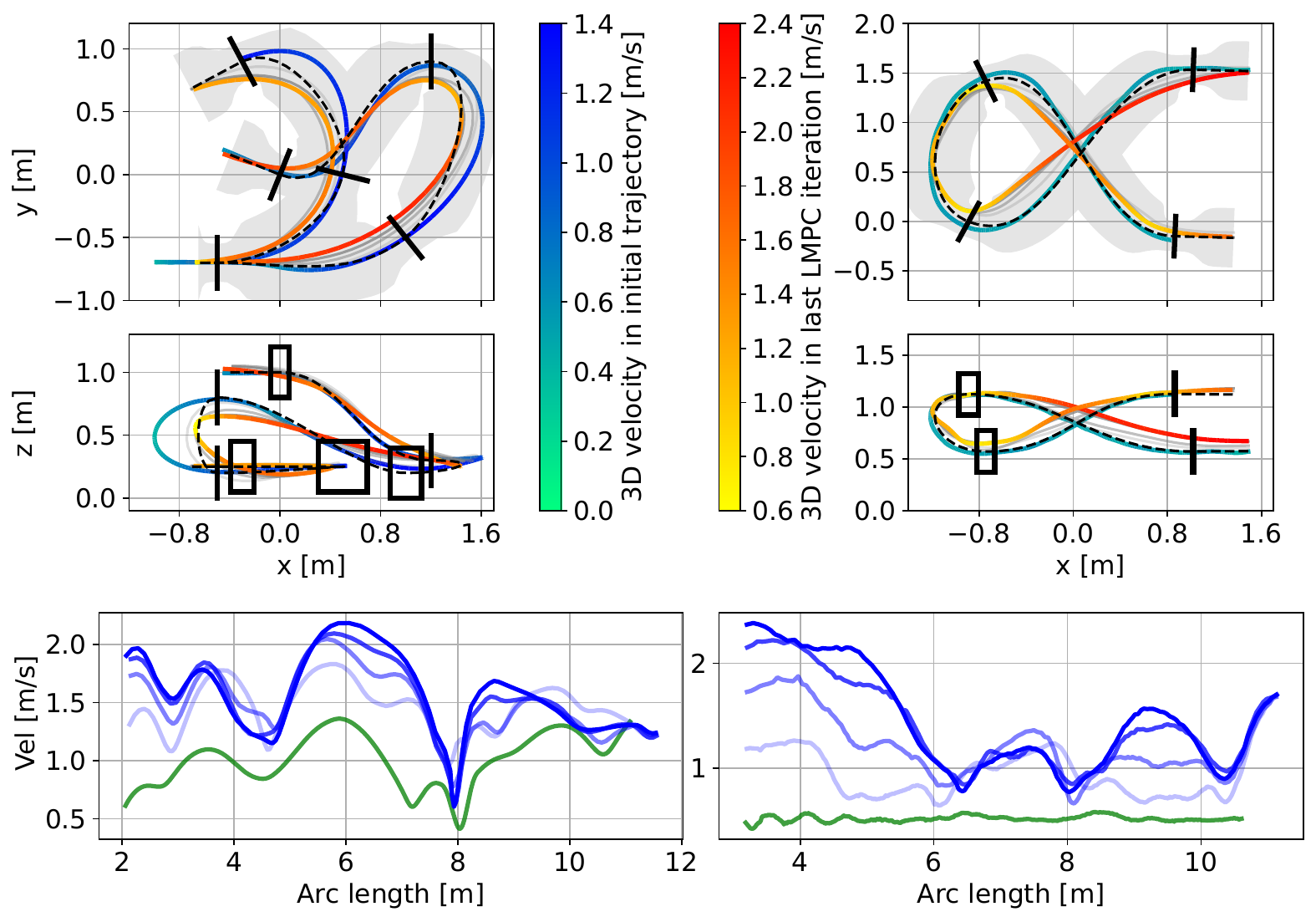}
\caption{The LMPC can iteratively improve the performance based on an initial demonstration in the simulation~(left) and the real-world~(right) on different tracks. In the simulation, the proposed LMPC is initialized by MPCC++ with $\mu=0.02$, and the real-world experiment is initialized by a PID controller~(see \autoref{fig:long_exposure}). The velocity profile for the initialization is marked in green, and the velocity profiles from the LMPC iterations are in blue. The brightness of the LMPC velocity profile increases with iterations.}
\label{fig:sim_real}
\end{figure}

We evaluate the performance of our algorithm using diverse controllers and parameter tunings in both simulation and the real world. As shown in Figure~\ref{fig:sim_real}, the simulation and real-world experiments are conducted on different tracks: a Split-S layout~\cite{song2023reaching} in simulation and a figure-eight track in the real world. All gates are square with a $\SI{0.4}{\meter}$ edge length. The statistics of the experimental results are presented in~\autoref{tab:sim_real}. We observe that the proposed LMPC controller can optimize initial trajectories generated by both controllers with all levels of tuning. Particularly in real-world experiments, the maximum improvement reached $60.85\%$, while even for aggressively tuned MPCC++, the average improvement over multiple trials still reached $6.05\%$. Importantly, the method does not require a predefined track structure; the central path is generated automatically from gate poses, enabling easy adaptation to unseen tracks. However, differences in converged lap times across controllers and tunings in simulations and real-world reveals that the final converged lap times are not identical. This indicates that the algorithm converges to local optima, primarily due to the nonlinearity and nonconvexity of the optimization problem. Therefore, initializing LMPC with a better trajectory can improve final performance after convergence.

\begin{table}[t]
\caption{LMPC performance initialized with different controllers and parameter-tunings in simulation and the real-world. The superscript $j_{\max}$ indicates the last iteration. The reported values represent the mean and standard deviation over eight independent trials.}
\label{tab:sim_real}
\centering
\begin{tabular}{c|c|c||c|c|c}
\cmidrule(lr){1-6}
\multirow{3}{*}{\textbf{Env}} 
& \multirow{3}{*}{\makecell{\textbf{Initial} \\ \textbf{Policy}}}
& \multirow{3}{*}{\makecell{\textbf{Hyper-} \\ \textbf{param.}}}
& \multicolumn{3}{c}{\textbf{Performance}}\\
\cmidrule(lr){4-6}
&&& \multicolumn{1}{c}{\textbf{Init.}} & \multicolumn{1}{c}{\textbf{Last.}} & \multicolumn{1}{c}{\textbf{Improve.}}\\
&&& {$t^0_{N_g}$ [\si{\second}]} & {$t^{j_{\max}}_{N_g}$ [\si{\second}]} & {$\Delta{t}/t^0_{N_g}$} \\
\cmidrule(lr){1-6}
\multirow{4}{*}{Sim}
& PID & $\lVert \vec{v}\rVert=0.5$ & $23.55 \pm 0.03$  & $8.42 \pm 0.13$ & \textbf{64.25\%} \\
\cmidrule(lr){2-6}
& \multirow{3}{*}{\makecell{MPCC \\ ++}}  & $\mu=0.02$ & $11.84\pm0.02$ & $6.04\pm0.06$ &\textbf{48.99\%} \\
&  & $\mu=0.05$ & $9.14\pm0.05$ & $6.11\pm0.05$ &\textbf{33.15\%} \\
&  & $\mu=0.10$ & $7.71\pm0.04$ & $5.92\pm0.06$ &\textbf{23.22\%} \\
\cmidrule(lr){1-6}
\multirow{4}{*}{Real} 
& PID & $\lVert \vec{v}\rVert=0.5$ & $17.09 \pm 0.11$ & $6.69 \pm 0.22$ &\textbf{60.85\%} \\
\cmidrule(lr){2-6}
& \multirow{3}{*}{\makecell{MPCC \\ ++}} & $\mu=0.02$ & $10.79 \pm 0.27$ & $7.51 \pm 0.38$ &\textbf{30.40\%} \\
& & $\mu=0.05$ & $7.62 \pm 0.11$ & $6.97 \pm 0.17$ &\textbf{~8.53\%} \\
& & $\mu=0.10$ & $6.45 \pm 0.12$ & $6.06 \pm 0.25$ &\textbf{~6.05\%} \\
\cmidrule(lr){1-6}
\end{tabular}
\label{Statistics of real drone experiment results over iterations}
\end{table}


\section{CONCLUSIONS}

In this paper, we augment the LMPC framework with an adaptive cost design and a modified local safe set to extend the original LMPC formulation to the drone racing scenario. Based on simulations, we analyzed the iterative performance mechanism of LMPC, identified the gap in applying the original LMPC to drone racing, and conducted ablation studies to validate the effectiveness of our proposed improvements. Extensive experiments have also been conducted to evaluate the practicality of our proposed algorithm. We have demonstrated that our improved algorithm can continuously optimize initial trajectories generated by any controller with any level of tuning in both simulation and real-world experiments, leading to improved racing performance. 






\bibliographystyle{IEEEtran}

\bibliography{ref}

\begin{thebibliography}{10}
\providecommand{\url}[1]{#1}
\csname url@samestyle\endcsname
\providecommand{\newblock}{\relax}
\providecommand{\bibinfo}[2]{#2}
\providecommand{\BIBentrySTDinterwordspacing}{\spaceskip=0pt\relax}
\providecommand{\BIBentryALTinterwordstretchfactor}{4}
\providecommand{\BIBentryALTinterwordspacing}{\spaceskip=\fontdimen2\font plus
\BIBentryALTinterwordstretchfactor\fontdimen3\font minus \fontdimen4\font\relax}
\providecommand{\BIBforeignlanguage}[2]{{%
\expandafter\ifx\csname l@#1\endcsname\relax
\typeout{** WARNING: IEEEtran.bst: No hyphenation pattern has been}%
\typeout{** loaded for the language `#1'. Using the pattern for}%
\typeout{** the default language instead.}%
\else
\language=\csname l@#1\endcsname
\fi
#2}}
\providecommand{\BIBdecl}{\relax}
\BIBdecl

\bibitem{shakhatreh2019unmanned}
H.~Shakhatreh, A.~H. Sawalmeh, A.~Al-Fuqaha, Z.~Dou, E.~Almaita, I.~Khalil, N.~S. Othman, A.~Khreishah, and M.~Guizani, ``Unmanned aerial vehicles ({UAVs}): A survey on civil applications and key research challenges,'' \emph{IEEE Access}, vol.~7, pp. 48\,572--48\,634, 2019.

\bibitem{lyu2023unmanned}
M.~Lyu, Y.~Zhao, C.~Huang, and H.~Huang, ``Unmanned aerial vehicles for search and rescue: A survey,'' \emph{Remote Sensing}, vol.~15, no.~13, p. 3266, 2023.

\bibitem{rejeb2023drones}
A.~Rejeb, K.~Rejeb, S.~J. Simske, and H.~Treiblmaier, ``Drones for supply chain management and logistics: a review and research agenda,'' \emph{International Journal of Logistics Research and Applications}, vol.~26, no.~6, pp. 708--731, 2023.

\bibitem{moon2019challenges}
H.~Moon, J.~Martinez-Carranza, T.~Cieslewski, M.~Faessler, D.~Falanga, A.~Simovic, D.~Scaramuzza, S.~Li, M.~Ozo, C.~De~Wagter \emph{et~al.}, ``Challenges and implemented technologies used in autonomous drone racing,'' \emph{Intelligent Service Robotics}, vol.~12, pp. 137--148, 2019.

\bibitem{foehn2022alphapilot}
P.~Foehn, D.~Brescianini, E.~Kaufmann, T.~Cieslewski, M.~Gehrig, M.~Muglikar, and D.~Scaramuzza, ``Alphapilot: Autonomous drone racing,'' \emph{Autonomous Robots}, vol.~46, no.~1, pp. 307--320, 2022.

\bibitem{hanover2024autonomous}
D.~Hanover, A.~Loquercio, L.~Bauersfeld, A.~Romero, R.~Penicka, Y.~Song, G.~Cioffi, E.~Kaufmann, and D.~Scaramuzza, ``Autonomous drone racing: A survey,'' \emph{IEEE Transactions on Robotics}, 2024.

\bibitem{song2023reaching}
Y.~Song, A.~Romero, M.~M{\"u}ller, V.~Koltun, and D.~Scaramuzza, ``Reaching the limit in autonomous racing: Optimal control versus reinforcement learning,'' \emph{Science Robotics}, vol.~8, no.~82, p. eadg1462, 2023.

\bibitem{krinner2024mpcc++}
M.~Krinner, A.~Romero, L.~Bauersfeld, M.~Zeilinger, A.~Carron, and D.~Scaramuzza, ``{MPCC}++: Model predictive contouring control for time-optimal flight with safety constraints,'' in \emph{Robotics: Science and Systems Conference (RSS 2024)}, 2024.

\bibitem{brunke2022safe}
L.~Brunke, M.~Greeff, A.~W. Hall, Z.~Yuan, S.~Zhou, J.~Panerati, and A.~P. Schoellig, ``Safe learning in robotics: From learning-based control to safe reinforcement learning,'' \emph{Annual Review of Control, Robotics, and Autonomous Systems}, vol.~5, no.~1, pp. 411--444, 2022.

\bibitem{rosolia2017learning}
U.~Rosolia and F.~Borrelli, ``Learning model predictive control for iterative tasks. a data-driven control framework,'' \emph{IEEE Transactions on Automatic Control}, vol.~63, no.~7, pp. 1883--1896, 2017.

\bibitem{kamel2017linear}
M.~Kamel, M.~Burri, and R.~Siegwart, ``Linear vs nonlinear mpc for trajectory tracking applied to rotary wing micro aerial vehicles,'' \emph{IFAC-PapersOnLine}, vol.~50, no.~1, pp. 3463--3469, 2017.

\bibitem{sun2022comparative}
S.~Sun, A.~Romero, P.~Foehn, E.~Kaufmann, and D.~Scaramuzza, ``A comparative study of nonlinear mpc and differential-flatness-based control for quadrotor agile flight,'' \emph{IEEE Transactions on Robotics}, vol.~38, no.~6, pp. 3357--3373, 2022.

\bibitem{romero2022model}
A.~Romero, S.~Sun, P.~Foehn, and D.~Scaramuzza, ``Model predictive contouring control for time-optimal quadrotor flight,'' \emph{IEEE Transactions on Robotics}, vol.~38, no.~6, pp. 3340--3356, 2022.

\bibitem{kaufmann2023champion}
E.~Kaufmann, L.~Bauersfeld, A.~Loquercio, M.~M{\"u}ller, V.~Koltun, and D.~Scaramuzza, ``Champion-level drone racing using deep reinforcement learning,'' \emph{Nature}, vol. 620, no. 7976, pp. 982--987, 2023.

\bibitem{song2021autonomous}
Y.~Song, M.~Steinweg, E.~Kaufmann, and D.~Scaramuzza, ``Autonomous drone racing with deep reinforcement learning,'' in \emph{2021 IEEE/RSJ International Conference on Intelligent Robots and Systems (IROS)}, 2021, pp. 1205--1212.

\bibitem{BLANCHINI19991747}
F.~Blanchini, ``Set invariance in control,'' \emph{Automatica}, vol.~35, no.~11, pp. 1747--1767, 1999.

\bibitem{rosolia2017autonomous}
U.~Rosolia, A.~Carvalho, and F.~Borrelli, ``Autonomous racing using learning model predictive control,'' in \emph{2017 American Control Conference (ACC)}, 2017, pp. 5115--5120.

\bibitem{brunner2017repetitive}
M.~Brunner, U.~Rosolia, J.~Gonzales, and F.~Borrelli, ``Repetitive learning model predictive control: An autonomous racing example,'' in \emph{2017 IEEE 56th Annual Conference on Decision and Control (CDC)}, 2017, pp. 2545--2550.

\bibitem{werling2010optimal}
M.~Werling, J.~Ziegler, S.~Kammel, and S.~Thrun, ``Optimal trajectory generation for dynamic street scenarios in a frenét frame,'' in \emph{2010 IEEE International Conference on Robotics and Automation (ICRA)}, 2010, pp. 987--993.

\bibitem{li2022learning}
G.~Li, A.~Tunchez, and G.~Loianno, ``Learning model predictive control for quadrotors,'' in \emph{2022 IEEE International Conference on Robotics and Automation (ICRA)}, 2022, pp. 5872--5878.

\bibitem{verschueren2022acados}
R.~Verschueren, G.~Frison, D.~Kouzoupis, J.~Frey, N.~v. Duijkeren, A.~Zanelli, B.~Novoselnik, T.~Albin, R.~Quirynen, and M.~Diehl, ``acados—a modular open-source framework for fast embedded optimal control,'' \emph{Mathematical Programming Computation}, vol.~14, no.~1, pp. 147--183, 2022.

\bibitem{calogero2023learning}
L.~Calogero, M.~Mammarella, and F.~Dabbene, ``Learning model predictive control for quadrotors minimum-time flight in autonomous racing scenarios,'' \emph{IFAC-PapersOnLine}, vol.~56, no.~2, pp. 1063--1068, 2023.

\bibitem{Borrelli2017}
F.~Borrelli, A.~Bemporad, and M.~Morari, \emph{Predictive Control for Linear and Hybrid Systems}.\hskip 1em plus 0.5em minus 0.4em\relax Cambridge University Press, 2017.

\bibitem{QIN2003733}
S.~Qin and T.~A. Badgwell, ``A survey of industrial model predictive control technology,'' \emph{Control Engineering Practice}, vol.~11, no.~7, pp. 733--764, 2003.

\bibitem{rosolia2019learning}
U.~Rosolia and F.~Borrelli, ``Learning how to autonomously race a car: a predictive control approach,'' \emph{IEEE Transactions on Control Systems Technology}, vol.~28, no.~6, pp. 2713--2719, 2019.

\bibitem{Lukas2018}
L.~Brunke, ``Learning model predictive control for competitive autonomous racing,'' Master's thesis, Hamburg University of Technology, 2018.

\end{thebibliography}

\end{document}